%% file: MICCAI2026_main_conference_paper_template.tex
\documentclass[runningheads]{llncs}
\usepackage[T1]{fontenc}
\usepackage{marvosym}
\usepackage{graphicx,verbatim}
\usepackage{hyperref} 
\usepackage{cite}
\usepackage{amsmath}
\usepackage{placeins}
\usepackage{array}
\usepackage{booktabs}
\usepackage{xcolor}
\usepackage{multirow}

\newcolumntype{P}[1]{>{\centering\arraybackslash}p{#1}} 
\newcolumntype{R}[1]{>{\raggedleft\arraybackslash}p{#1}} 
\newcolumntype{L}[1]{>{\raggedright\arraybackslash}p{#1}}

\newlength\savewidth
 
\definecolor{lightblue}{RGB}{41,128,185}
\definecolor{myorange}{RGB}{255,140,0} 
\definecolor{darkorange}{RGB}{204,85,0} 
\definecolor{midorange}{RGB}{230,110,0} 

\usepackage{color}

\usepackage{makecell}

\newcommand{\msinline}[2]{%
  #1\raisebox{-0.5ex}{$\pm$#2}%
}

\begin{document}
\title{DUET: Dual-Paradigm Adaptive Expert Triage with Single-cell Inductive Prior for Spatial Transcriptomics Prediction}
\titlerunning{DUET}
%

\author{Junchao Zhu\inst{1}, Ruining Deng\inst{2}, Junlin Guo\inst{1}, Tianyuan Yao\inst{1}, Chongyu Qu\inst{1}, Juming Xiong\inst{1}, Zhengyi Lu\inst{1}, Yanfan Zhu\inst{1}, Marilyn Lionts\inst{1}, Yuechen Yang\inst{1}, Yu Wang\inst{3}, Shilin Zhao\inst{3}, Haichun Yang\inst{3}, Yuankai Huo\inst{1}\thanks{Email: \href{email:yuankai.huo@vanderbilt.edu}{yuankai.huo@vanderbilt.edu}}}  
\authorrunning{Zhu et al.}

\institute{Vanderbilt University, Tennessee, USA \\
\and Weill Cornell Medicine, New York, USA \\ 
\and Vanderbilt University Medical Center, Tennessee, USA \\}

\maketitle              
\input{Sec/0_abstract}
\input{Sec/1_introduction}
\input{Sec/2_method}

\input{Sec/3_experiments}
\input{Sec/4_results}
\input{Sec/5_conclusion}

    



%
%
%
\bibliographystyle{unsrt}
\bibliography{mybibliography}

\end{document}

%% file: Sec/0_abstract.tex
\begin{abstract}
Inferring spatially resolved gene expression from histology images offers a cost-effective complement to spatial transcriptomics (ST). However, existing methods reduce this task to a simple morphology-to-expression mapping, where visual similarity does not guarantee molecular consistency. Meanwhile, single-cell data has amassed rich resources far surpassing the scale of ST data, yet it remains underexplored in vision-omics modeling. Furthermore, current approaches commit to a monolithic paradigm with bottlenecks, unable to balance expressive flexibility with biological fidelity. To bridge these gaps,  we propose DUET, a novel dual-paradigm framework that synergizes parametric prediction and memory-based retrieval under cellular inductive priors. DUET implements a parallel regression-retrieval paradigm, adaptively reconciling the outputs of its complementary pathways. To mitigate aleatoric vision ambiguity, we incorporate large-scale single-cell references to impose molecular states as biological constraints for faithful learning. Building upon structural refinement, we further design a lightweight adapter to dynamically assign branch preference across spatial contexts to achieve optimal performance. Extensive experiments on three public datasets across varied gene scales demonstrate that DUET achieves SOTA performance, with consistent gains contributed by each proposed component. Code is available at \href{https://github.com/Junchao-Zhu/DUET}{https://github.com/Junchao-Zhu/DUET}.

\keywords{Computational Pathology  \and Spatial Transcriptomics \and  scRNA-sequencing Genomics.}

\end{abstract}

%% file: Sec/1_introduction.tex
\section{Introduction}
\begin{figure*}[htbp]
    \centering
\includegraphics[width=0.98\textwidth]{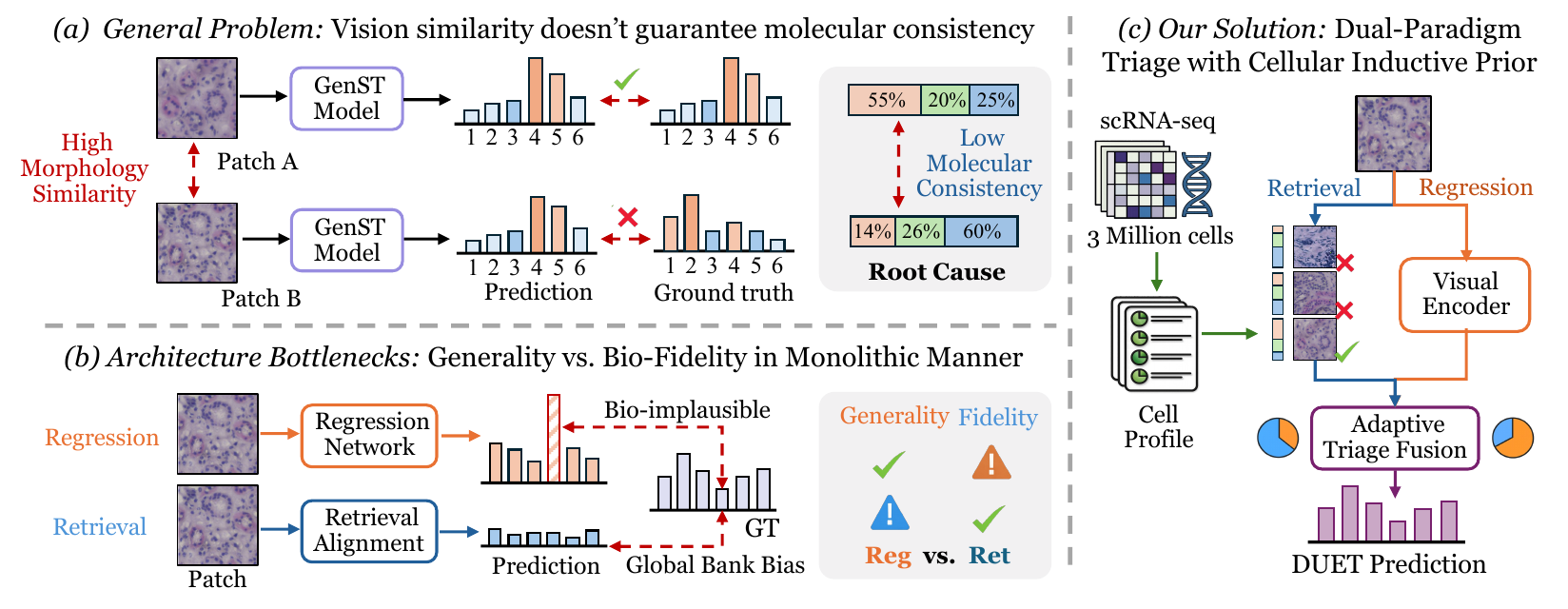}
    \caption{\textbf{Paradigm limitations in vision-omics modeling.} (a) Morphological visual similarity does not guarantee molecular consistency. (b) Monolithic learning manner fails to reconcile flexibility with biological reliability. (c) Proposed DUET introduces a dual-paradigm framework with cellular priors and adaptive fusion, jointly modeling parametric prediction and memory-based inference for biologically reliable estimation. }
    \label{fig:problem_definition}
\end{figure*}
Spatial transcriptomics (ST) links tissue morphology with spatial molecular programs~\cite{burgess2019spatial, asp2019spatiotemporal,asp2020spatially}. However, its high cost constrains data scale, while sequencing noise and insufficient read depth further weaken the available supervisory signal~\cite{choe2023advances, jin2024advances}. Against this backdrop, deep learning methods~\cite{ke2023clusterseg,madhu2025heist, qu2025post,zhu2023anti, huang2025stpath} have brought a new view in medical image analysis. These models can capture complex visual patterns and nonlinear relationships between tissue structure and molecular states~\cite{he2020integrating, yang2023exemplar, xie2024spatially}. Inferring spatial expression profiles from histology images has offered a cost-effective strategy~\cite{he2020integrating, zhu2025asign, yang2025staig}. 
Yet, existing methods simplify the task into a vision-to-expression mapping~\cite{xie2024spatially, yang2023exemplar, zhu2025magnet}; however, high visual similarity does not guarantee molecular consistency, leading to aleatoric morphology ambiguity in molecular modeling, as presented in Fig.~\ref{fig:problem_definition}a. Meanwhile, single-cell RNA sequencing (scRNA-seq) has amassed cell-type and state references far beyond the scale of ST~\cite{regev2017human}, but remains underutilized in vision-omics modeling.

Moreover, current vision-driven methods commit to a monolithic paradigm~\cite{zhu2025computer,min2024multimodal,pang2021leveraging,zhu2025img2st}. As shown in Fig.~\ref{fig:problem_definition}b, regression-based models~\cite{yang2023exemplar,he2020integrating, han2025towards} provide expressive flexibility through parametric generalization, yet may yield biologically implausible outputs. Retrieval-based methods~\cite{chen2025visual, xie2024spatially, qu2026cohort,zhu2025scr2}, by anchoring predictions to reference expression distributions, ensure biological reliability but limit extrapolation beyond the empirical support.  Monolithic learning manner has each bottleneck, while an adaptive fusion of both complementary paradigms could trade off expressive freedom and biological fidelity across spatial contexts.

To achieve this, we introduce DUET, a novel dual-paradigm framework that integrates cellular inductive priors and adaptive expert triage to synergize flexibility and biological reliability. DUET establishes a coordinated regression-retrieval architecture, where parametric prediction and memory-based inference operate in parallel and are adaptively fused, as illustrated in Fig.~\ref{fig:framework}. Moreover, we leverage large-scale scRNA-seq data to infer spot cell-type composition, incorporating cellular inductive signals to filter out biologically incompatible neighbors and reinforce coherence. Building upon structure-aware refinement, we further introduce a lightweight adapter trained on a small held-out set to perform expert triage, dynamically learning fusion weights that assign branch preference for each spatial location. Our contributions can be summarized as threefold:

\begin{itemize}
\item  We propose a pioneering dual-paradigm framework DUET that transcends monolithic modeling by unifying parametric prediction and memory-based inference to reconcile flexibility and biological fidelity.

\item We curate large-scale scRNA-seq references to impose cellular inductive priors as biological constraints, alleviating morphological ambiguity with plausible molecular states for faithful vision-omics modeling.
\begin{figure*}[htbp]
    \centering
\includegraphics[width=1\textwidth]{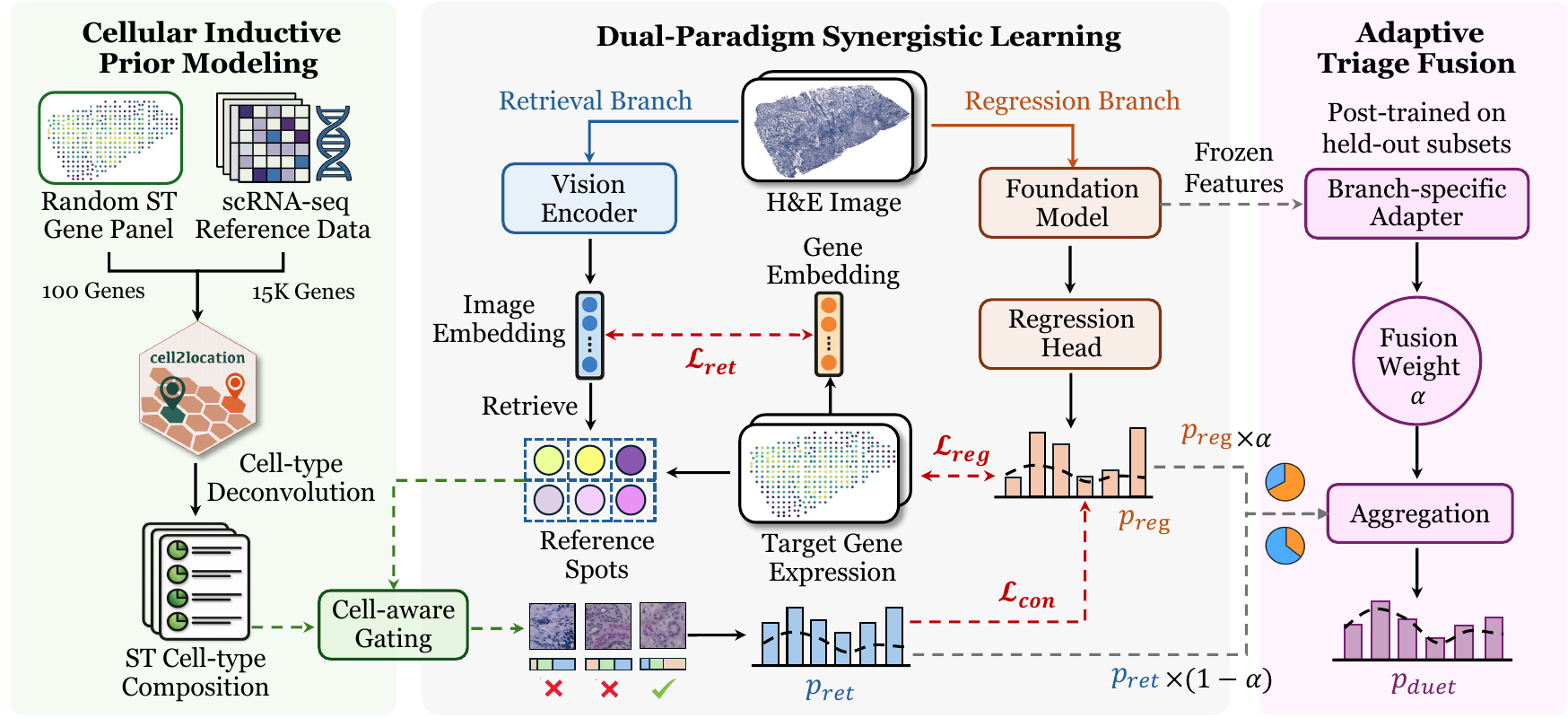}
    \caption{\textbf{Overview of proposed DUET framework.} DUET comprises three key components. We first leverage large-scale scRNA-seq references to perform cell-type deconvolution and derive cellular inductive priors, enabling a gating mechanism to filter biologically incompatible candidates. We then conduct unified dual-paradigm joint training and facilitate cross-branch knowledge transfer through a retrieval-guided consistency loss. Finally, a lightweight adapter is introduced for adaptive triage fusion, dynamically routing each spot to its more reliable paradigm for the final output. }
    \label{fig:framework}
\end{figure*}
\item We design an adaptive expert triage module equipped with a lightweight adapter to dynamically modulate the contribution of regression and retrieval pathways for optimal expression prediction.

\end{itemize}

%% file: Sec/2_method.tex
\section{Method}
\subsection{Cellular Inductive Prior Modeling}\label{sec:prior}
Expression profile of ST spot consists of mixed signals from multiple cell types~\cite{gaspard2025cell}. Accurate estimation of cell-type composition is therefore essential for reliable modeling. To address this, we construct cellular inductive priors by integrating large-scale single-cell references. Given multi-batch scRNA-seq datasets $\mathcal{D}_{\text{sc}} = \{(\mathbf{x}_c, t_c, b_c)\}_{c=1}^{C}$,  where $\mathbf{x}_c \in \mathcal{Z}_{\geq 0}^{G}$ denotes the raw UMI counts of cell $c$, 
$t_c \in \{1, \dots, T\}$ represents cell type, and $b_c$ indicates batch label. We adopt a negative binomial regression model~\cite{kleshchevnikov2022cell2location} to learn batch-robust cell-type signatures:
\begin{equation}
x_{cg} \sim \mathrm{NB}(\mu_{cg}, \theta_g), 
\quad 
\mu_{cg} = l_c \cdot \exp\!\left(m_g^{(b_c)}\right) \cdot \mu_{t_c,g}
\end{equation}
where $l_c$ denotes scaling factor, $m_g^{(b_c)}$ represents batch effect,  and $\mu_{t,g}$ is the average expression of gene $g$ in cell type $t$. After optimization via variational inference, we obtain the signature matrix $\mathbf{M} \in \mathcal{R}_{>0}^{G \times T}$,  where $M_{gt} = \hat{\mu}_{t,g}$ encodes the characteristic pattern of cell type.
Based on the signature matrix $\mathbf{M}$, we then perform cell-type deconvolution on ST data. Specifically, we randomly sample 100 genes non-overlapped with target genes to construct a deconvolution gene panel $G_{\text{de}}$. 
We then conduct Bayesian deconvolution~\cite{kleshchevnikov2022cell2location} for each ST slide:
\begin{equation}
y_{sg} \sim \mathrm{NB}\!\left(\mu_{sg}, \alpha_g\right), 
\quad 
\mu_{sg} = d_s \cdot \sum_{t=1}^{T} w_{st} \cdot M_{gt}
\end{equation}
where $w_{st} \geq 0$ denotes the abundance of cell type $t$ in spot $s$, and $d_s$ represents detection efficiency. The 5\% quantile $\hat{w}_{st}^{\mathrm{q05}}$ is then extracted and normalized. We then apply CellViT~\cite{horst2024cellvit} to perform cell count $n_s$ for each spot to construct a cellular gating signal as $g_{st} = n_s \cdot \hat{w}_{st}^{\mathrm{q05}}$ for the following learning.

\subsection{Dual-Paradigm Synergistic Learning}
\noindent \textbf{Cross-Modality Alignment.} We adopt DenseNet-121~\cite{huang2017densely} as visual encoder, followed by a projection head that maps and $\ell_2$-normalizes image features into embeddings $v_i \in \mathcal{R}^d$. A gene encoder with similar projects expression vector $y_i$ into $h_i \in \mathcal{R}^d$. Following~\cite{xie2024spatially}, we align these two modalities via a symmetric InfoNCE loss $\mathcal{L}_{\mathrm{ret}}$ with temperature $\tau=0.07$, defined as:
\begin{equation}
\mathcal{L}_{\mathrm{ret}} = -\frac{1}{2N}\sum_{i=1}^{N}\!\left[\log\frac{\exp(v_i^\top h_i / \tau)}{\sum_{j=1}^{N}\exp(v_i^\top h_j / \tau)} + \log\frac{\exp(h_i^\top v_i / \tau)}{\sum_{j=1}^{N}\exp(h_i^\top v_j / \tau)}\right]
\end{equation}

\noindent \textbf{Cell-Aware Gating Retrieval.} After training, we encode all training expression vectors into an embedding database $\{h_j\}_{j=1}^{N}$. For a query spot $s$, we compute similarity $\phi_{sj} = v_s^\top h_j$ and select top-$150$ entries to construct candidate set. However, relying solely on embedding similarity may introduce false-positive references with mismatched molecular compositions. We leverage the gating signal $g_s$  to perform cell-aware filtering on candidates. Let $\mathrm{sim}_{sj} = \cos(g_s,\,g_j)$ composition similarity. For query spot $s$ and candidate spot $j$, we retain only those candidates that satisfy cell count deviation and composition similarity constraints:
\begin{equation}
m_{sj} = 1\!\left[\frac{\left|\sum_t g_{st} - \sum_t g_{jt}\right|}{\max\!\left(\sum_t g_{st},\, \sum_t g_{jt}\right)} \leq \tau_c \right] \cdot 1\!\left[\mathrm{sim}_{sj} \geq \tau_p\right]
\end{equation}
where $\tau_c{=}0.5$ and $\tau_p{=}0.3$ are cell count deviation and composition similarity threshold, respectively. Then the final retrieval score is defined as :
\begin{equation}
r_{sj} = (1{-}\beta)\,\phi_{sj} + \beta\,\mathrm{sim}_{sj}, \quad \text{s.t.}\; m_{sj}{=}1
\end{equation}
where $\beta=0.3$ controls the trade-off between the two similarities. We rank candidates by $r_{sj}$ and select top-$k$ (or all remaining). The retrieval prediction $p_{ret}$ is obtained as a softmax-weighted average of the retrieved expression vectors.

\noindent \textbf{Regression and Dynamic Soft Consistency.}
To enhance our framework's extrapolation capability, we employ the pathology foundation model CONCH~\cite{lu2024visual} to extract features $f_i$, which are then passed through a multi-layer regression head to produce $p_{reg}$. However, under limited samples and high-dimensional output, purely supervised regression may produce biologically implausible predictions. To reduce estimation uncertainty, we introduce the structured predictions from retrieval as a soft constraint $\mathcal{L}_{\text{con}}=\lambda(e) \cdot \mathrm{MSE}(p_{reg},\,p_{ret})$. The joint optimization objective of the regression branch is formulated as:
\begin{equation}
\mathcal{L}_{\text{reg}} = \mathrm{MSE}(p_{reg},\,y) + \lambda(e) \cdot \mathrm{MSE}(p_{reg},\,p_{ret})
\end{equation}
where guidance weight $\lambda(e)=\tfrac{\lambda_0}{2}(1+\cos\tfrac{\pi e}{E_d})$ cosine-anneals from $\lambda_0=1$ to 0 over first $E_d=30$ epochs. Thereby, it constrains the output to the biological expression structure in retrieval at early stage, while this constraint is released to allow free optimization later. Embedding database $\{h_j\}_{j=1}^{N}$ will be fresh each epoch to ensure soft guidance signal improves in tandem with embedding quality.

\subsection{Adaptive Expert Triage Fusion}
We design a lightweight adaptive fusion module to learn spot-level preference for route predictions. We first take frozen features $f_s \in \mathcal{R}^{D}$ from the vision encoder in the regression branch and use them as input to a lightweight MLP that predicts a spot-level fusion weight $\alpha_s$. Specifically, the fusion weight is defined as: $\alpha_s = \tfrac{1}{2} + \tfrac{1}{2}\tanh\!\big(\mathrm{MLP}(f_s)\big)$, where the $\tanh$ function naturally constrains $\alpha_s \in (0,1)$. The final fused prediction $y_{duet} $ is then formulated as:
\begin{equation}
y_{duet} = \alpha_s \cdot y_{ret} + (1 - \alpha_s) \cdot y_{reg}
\end{equation}

This module is trained in a post-hoc stage with both branches fully frozen, using a small set of held-out slides from the training set.  The optimization objective is defined as $\mathcal{L}_{fuse} = \mathrm{MSE}(y_{duet},\,y) + \|\boldsymbol{\delta}\|_2^2$, where $\boldsymbol{\delta} = \alpha_s - 0.5$ is the residual deviation from equal weighting, and the regularization term $\|\boldsymbol{\delta}\|_2^2$ prevents overfitting on the small-scale held-out data, enabling the adapter to dynamically modulate pathway contributions for each spot and yield final estimation of $y_{duet}$.

%% file: Sec/3_experiments.tex
\section{Data and Experiments}

\noindent \textbf{Datasets and Preprocessing.}
All methods are fairly assessed on three publicly available ST datasets, including HER2~\cite{andersson2021spatial}, Breast Cancer~\cite{he2020integrating}, and Kidney~\cite{lake2023atlas}. We extract a $224 \times 224$ image patch centered at each spot's coordinates as the model input. As prediction targets, we consider the top 100, 300, and 500 high-variance genes (HVGs). Consistent with BLEEP~\cite{xie2024spatially}, raw gene expression counts are transformed using $\log(1+x)$. For external single-cell references, we adopt two million cells from~\cite{lake2025cellular} for the Kidney dataset, and approximately three million cells collected from~\cite{chen2025highly, reed2024single, klughammer2024multi} for the Breast Cancer and HER2 datasets.

\noindent \textbf{Baseline and Evaluation Metrics.} We benchmark our DUET against SOTA methods, covering regression-based ST-Net~\cite{he2020integrating}, EGN~\cite{yang2023exemplar} and His2ST~\cite{zeng2022spatial}; retrieval-based  BLEEP~\cite{xie2024spatially} and mclSTExp~\cite{min2024multimodal}; and foundation models OmiCLIP~\cite{chen2025visual} and UMPIRE~\cite{han2025towards}. All methods were implemented under the same protocols. Model performance was evaluated using Pearson correlation coefficient (PCC), mean squared error (MSE), and mean absolute error (MAE).

\noindent \textbf{Implementation Details.}
All experiments were performed on two NVIDIA RTX A6000 GPUs. We adopted SGD optimizer with a momentum of 0.9 and a weight decay of $10^{-4}$. The learning rate was set to $lr_0 = 10^{-4}$, and the batch size was set to 128. For retrieval, we set $\tau_c = 0.5$, $\tau_p = 0.15$, and $\text{top-}k = 100$.

\begin{table*}[t]
\centering
\caption{\textbf{Quantitative comparisons on expression prediction. } Best performance is highlighted in \textcolor{midorange}{\underline{orange}}, where DUET outperforms SOTAs across datasets and HVGs.}
\label{tab:gene_prediction}
{%
  \fontsize{8pt}{\dimexpr0.93\baselineskip}\selectfont
  \begin{tabular}{
      p{0.055\linewidth}
      p{0.155\linewidth}
      P{0.0777\linewidth} P{0.0777\linewidth} P{0.0777\linewidth}
      P{0.0777\linewidth} P{0.0777\linewidth} P{0.0777\linewidth}
      P{0.0777\linewidth} P{0.0777\linewidth} P{0.0777\linewidth}
  }
  \toprule
  \multirow{2}{*}{HVG} & \multirow{2}{*}{Model} &
  \multicolumn{3}{c}{Breast Cancer~\cite{he2020integrating}} &
  \multicolumn{3}{c}{HER2~\cite{andersson2021spatial}}&
  \multicolumn{3}{c}{Kidney~\cite{lake2023atlas}} \\
  \cline{3-11}
   & & MSE $\downarrow$ & MAE $\downarrow$ & PCC $\uparrow$
   & MSE $\downarrow$ & MAE $\downarrow$ & PCC $\uparrow$
   & MSE $\downarrow$ & MAE $\downarrow$ & PCC $\uparrow$ \\
  \midrule

  \multirow{8}{*}{\centering 100}
   & ST-Net~\cite{he2020integrating}    & 0.4126 & 0.5179 & 0.1270 & 0.6083 & 0.6157 & 0.2260 & 0.5412 & 0.5925 & 0.1672 \\
   & His2ST~\cite{zeng2022spatial}    & 0.4517 & 0.5525 & 0.0822 & 0.6533 & 0.6588 & 0.1413 & 0.5231 & 0.5879 & 0.0778 \\
   & EGN~\cite{yang2023exemplar}       & 0.4294 & 0.5297 & 0.1216 & 0.5866 & 0.6031 & 0.2312 & 0.5179 & 0.5291 & 0.1182 \\
   & BLEEP~\cite{xie2024spatially}     & 0.3738 & 0.4794 & 0.1975 & 0.6810 & 0.6183 & 0.2223 & 0.5923 & 0.6028 & 0.1898 \\
   & mclSTExp~\cite{min2024multimodal}  & 0.5141 & 0.5333 & 0.0950 & 0.6467 & 0.6125 & 0.2168 & 0.5804 & 0.6026 & 0.1942 \\
   & OmiCLIP~\cite{chen2025visual}   & 0.3945 & 0.5006 & 0.2432 & 0.6033 & 0.6028 & 0.2649 & 0.4865 & 0.5639 & 0.2227 \\
   & UMPIRE~\cite{han2025towards}    & 0.4209 & 0.5213 & 0.1905 & 0.5802 & 0.6028 & 0.2385 & 0.4529 & 0.5379 & 0.2496 \\
   & \underline{DUET (Ours)}      & \textcolor{midorange}{\underline{0.2886}} & \textcolor{midorange}{\underline{0.4117}} & \textcolor{midorange}{\underline{0.3583}}
                 & \textcolor{midorange}{\underline{0.5726}} & \textcolor{midorange}{\underline{0.5906}} & \textcolor{midorange}{\underline{0.3038}}
                 & \textcolor{midorange}{\underline{0.4210}} & \textcolor{midorange}{\underline{0.5286}} & \textcolor{midorange}{\underline{0.3099}} \\

  \midrule

  \multirow{8}{*}{\centering 300}
   & ST-Net~\cite{he2020integrating}    & 0.4745 & 0.5525 & 0.1133 & 0.6201 & 0.6018 & 0.2240 & 0.5038 & 0.5899 & 0.1466 \\
   & His2ST~\cite{zeng2022spatial}    & 0.4399 & 0.5489 & 0.0501 & 0.6345 & 0.6496 & 0.1339 & 0.5212 & 0.5889 & 0.0539 \\
   & EGN~\cite{yang2023exemplar}       & 0.4164 & 0.5277 & 0.1214 & 0.6771 & 0.6809 & 0.1549 & 0.5566 & 0.6047 & 0.1061 \\
   & BLEEP~\cite{xie2024spatially}     & 0.3593 & 0.4778 & 0.1860 & 0.6311 & 0.5968 & 0.2195 & 0.5428 & 0.5481 & 0.2473 \\
   & mclSTExp~\cite{min2024multimodal}  & 0.4307 & 0.5046 & 0.1142 & 0.6661 & 0.6189 & 0.1958 & 0.6412 & 0.6247 & 0.1736 \\
   & OmiCLIP~\cite{chen2025visual}   & 0.3903 & 0.5045 & 0.1843 & 0.5672 & 0.5884 & 0.2382 & 0.4710 & 0.5576 & 0.2127 \\
   & UMPIRE~\cite{han2025towards}    & 0.4208 & 0.5309 & 0.2169 & 0.5838 & 0.6104 & 0.2500 & 0.4586 & 0.5492 & 0.2748 \\
   & \underline{DUET (Ours)}      & \textcolor{midorange}{\underline{0.2959}} & \textcolor{midorange}{\underline{0.4321}} & \textcolor{midorange}{\underline{0.3082}}
                 & \textcolor{midorange}{\underline{0.5297}} & \textcolor{midorange}{\underline{0.5668}} & \textcolor{midorange}{\underline{0.3158}}
                 & \textcolor{midorange}{\underline{0.4067}} & \textcolor{midorange}{\underline{0.5196}} & \textcolor{midorange}{\underline{0.3105}} \\

  \midrule

  \multirow{8}{*}{\centering 500}
   & ST-Net~\cite{he2020integrating}    & 0.4849 & 0.5796 & 0.1094 & 0.6485 & 0.6459 & 0.1901 & 0.4957 & 0.5678 & 0.1359 \\
   & His2ST~\cite{zeng2022spatial}    & 0.4419 & 0.5496 & 0.0799 & 0.6359 & 0.6508 & 0.0769 & 0.5274 & 0.5881 & 0.0400 \\
   & EGN~\cite{yang2023exemplar}       & 0.4232 & 0.5323 & 0.1155 & 0.6519 & 0.6648 & 0.2135 & 0.4937 & 0.5833 & 0.0960 \\
   & BLEEP~\cite{xie2024spatially}     & 0.3539 & 0.4747 & 0.2255 & 0.6335 & 0.6042 & 0.2037 & 0.5262 & 0.5755 & 0.1930 \\
   & mclSTExp~\cite{min2024multimodal}  & 0.4073 & 0.4975 & 0.1461 & 0.6416 & 0.6083 & 0.2100 & 0.5672 & 0.5939 & 0.1367 \\
   & OmiCLIP~\cite{chen2025visual}   & 0.4007 & 0.5139 & 0.2522 & 0.5540 & 0.5826 & 0.2551 & 0.4873 & 0.5632 & 0.2088 \\
   & UMPIRE~\cite{han2025towards}    & 0.4249 & 0.5336 & 0.1739 & 0.5745 & 0.6042 & 0.2416 & 0.4751 & 0.5565 & 0.2583 \\
   & \underline{DUET (Ours)}      & \textcolor{midorange}{\underline{0.3095}} & \textcolor{midorange}{\underline{0.4481}} & \textcolor{midorange}{\underline{0.3008}}
                 & \textcolor{midorange}{\underline{0.5248}} & \textcolor{midorange}{\underline{0.5654}} & \textcolor{midorange}{\underline{0.3150}}
                 & \textcolor{midorange}{\underline{0.4099}} & \textcolor{midorange}{\underline{0.5078}} & \textcolor{midorange}{\underline{0.3153}} \\

  \bottomrule
  \end{tabular}
}
\end{table*}

%% file: Sec/4_results.tex
\section{Results}

\noindent \textbf{Empirical Cross-validation on Expression Prediction.} We evaluate DUET against SOTAs using four-fold sample-level cross-validation. Table~\ref{tab:gene_prediction} summarizes quantitative comparisons across cohorts and HVG settings, where DUET consistently achieves best performance on all metrics. For HER2 with 300 HVGs, DUET reaches $0.3158\pm0.135$ of PCC and $0.5297\pm0.141$ of MSE, substantially outperforming single-manner models BLEEP ($0.2195\pm0.175$ / $0.6311\pm0.167$) and OmiCLIP ($0.2500\pm0.185$ / $0.5838\pm0.207$). This advantage is consistently maintained as the number of HVGs varies, indicating its stronger adaptability. 

\begin{figure*}[h]
    \centering
\includegraphics[width=\textwidth]{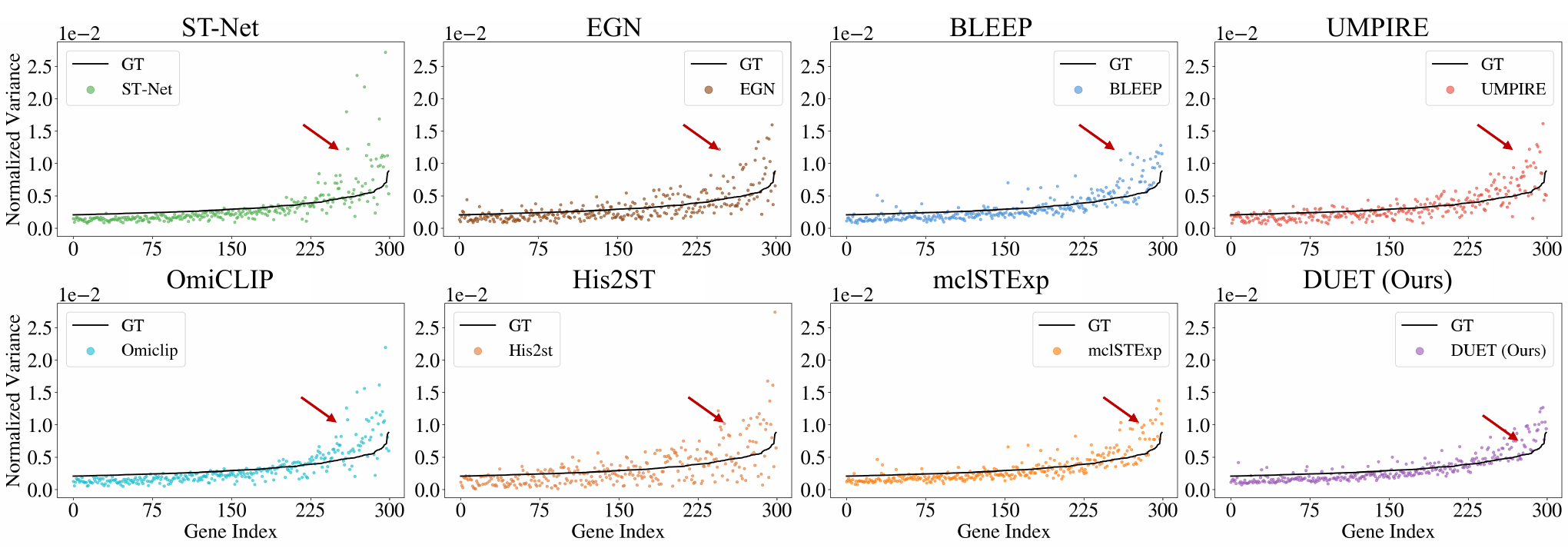}
    \caption{Normalized variance of predictions on Kidney (300 HVGs). Red arrows highlight the failure of current methods to preserve the bio-fidelity of inter-gene variance pattern.}
    \label{fig:var_visual}
\end{figure*}

Following~\cite{zhu2025diffusion}, we further evaluate prediction fidelity by computing the normalized variance and sorting in ascending order of truth variance (Fig.~\ref{fig:var_visual}). Regression-based methods (e.g., His2ST, UMPIRE) markedly overestimate highly variable genes, leading to unstable predictions. In contrast, DUET closely follows the ground-truth curve across the entire variance spectrum, avoiding over-smoothing in low-variance genes and noise amplification in high-variance ones. These results suggest that our dual-paradigm design effectively regularizes the output distribution while preserving bio-meaningful gene variability patterns.

\begin{figure*}[h]
    \centering
\includegraphics[width=\textwidth]{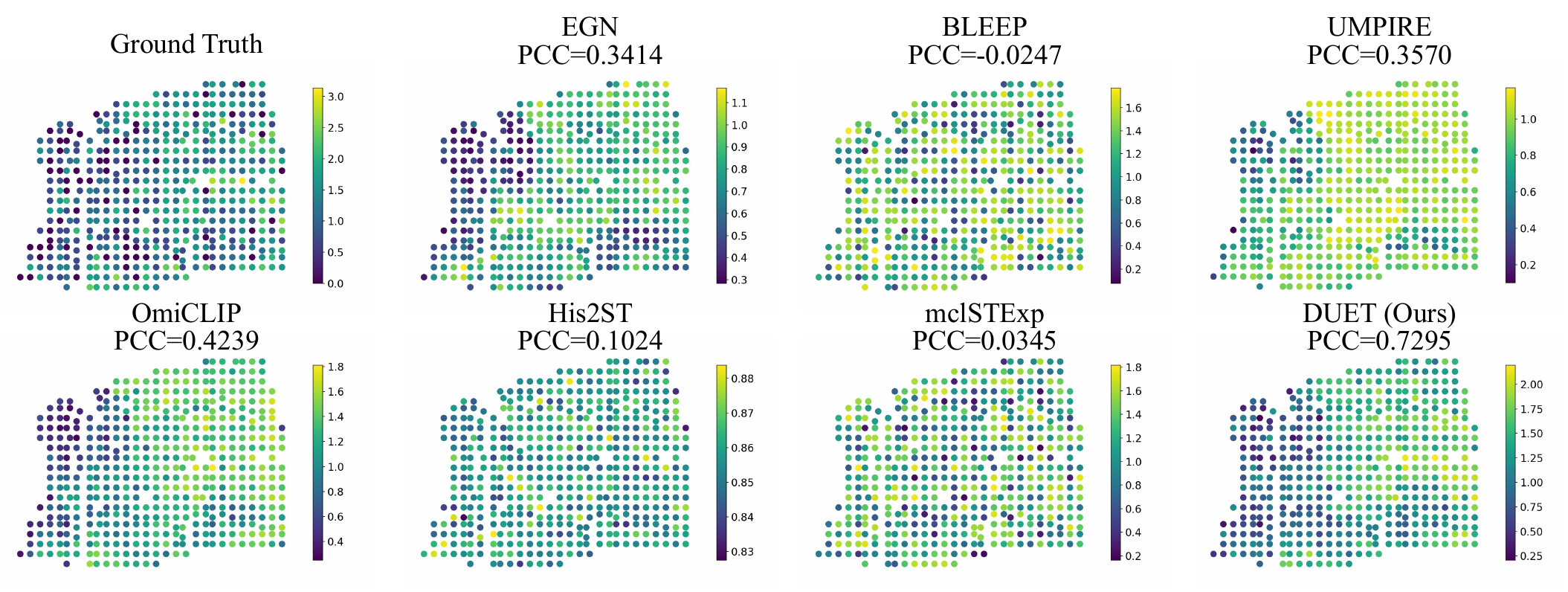}
    \caption{Predicted spatial expression distribution of breast-cancer-related gene ACTN4.}
    \label{fig:result}
\end{figure*}

\noindent \textbf{Pivotal Gene Expression Prediction.}
We identified $\alpha$-actinin-4 (ACTN4)~\cite{wang2017direct,tentler2019role}, a key breast-cancer–associated marker, to assess models' practical applicability. Fig. \ref{fig:result} visualizes the spatial pattern of ACTN4 along with the PCC across different models. Our DUET achieved the best PCC of 0.729. In contrast, monolithic-paradigm methods such as EGN (0.341) or BLEEP (-0.024) failed to reliably recover ACTN4 due to limited ability to capture cross-sample spatial heterogeneity. By incorporating a cellular prior and dual-paradigm adaptive design, DUET demonstrates strong utility in real-world clinical scenarios.

\begin{table*}[thbp]
\centering
\caption{\textbf{Ablation study for the functional blocks in DUET with 300 HVGs,} where best performance is highlighted in \textcolor{midorange}{\underline{orange}} and the second is highlighted in \textcolor{lightblue}{blue.}}
\label{tab:ablation_modules}
{%
  \fontsize{8pt}{\dimexpr0.93\baselineskip}\selectfont
  \begin{tabular}{
      p{0.145\linewidth}
      P{0.133\linewidth} P{0.133\linewidth} P{0.133\linewidth}
      P{0.133\linewidth} P{0.133\linewidth} P{0.133\linewidth}
  }
  \toprule
  \multirow{2}{*}{Setting} &
  \multicolumn{3}{c}{HER2~\cite{andersson2021spatial}} &
  \multicolumn{3}{c}{Kidney~\cite{lake2023atlas}} \\
  \cline{2-7}
   & MSE $\downarrow$ & MAE $\downarrow$ & PCC $\uparrow$
   & MSE $\downarrow$ & MAE $\downarrow$ & PCC $\uparrow$ \\
  \midrule

  w.o. $\mathcal{L}_{\text{con}}$
   & \msinline{0.589}{.192} & \msinline{0.608}{.084} & \msinline{0.255}{.175}
   & \msinline{0.471}{.045} & \msinline{0.556}{.022} & \msinline{0.267}{.024} \\

  \underline{DUET$_{Reg}$}
   & \msinline{0.574}{.184} & \msinline{0.581}{.104} & \msinline{0.281}{.147}
   & \msinline{0.478}{.031} & \msinline{0.549}{.013} & \msinline{0.273}{.037} \\

  \midrule

  w.o. Gating
   & \msinline{0.631}{.158} & \msinline{0.597}{.083} & \msinline{0.220}{.128}
   & \msinline{0.543}{.026} & \msinline{0.548}{.015} & \msinline{0.249}{.029} \\

  $\tau_c,\tau_p$=0.1,0.1
   & \msinline{0.598}{.118} & \msinline{0.589}{.066} & \msinline{0.239}{.133}
   & \msinline{0.481}{.036} & \msinline{0.533}{.019} & \msinline{0.251}{.022} \\

  $\tau_c,\tau_p$=0.7,0.5
   & \msinline{0.581}{.143} & \msinline{0.590}{.077} & \msinline{0.242}{.129}
   & \msinline{0.490}{.024} & \msinline{0.525}{.013} & \msinline{0.273}{.030} \\

  \underline{DUET$_{Ret}$}
   & \msinline{0.569}{.134} & \msinline{0.586}{.074} & \msinline{0.266}{.134}
   & \msinline{0.488}{.027} & \msinline{0.526}{.015} & \msinline{0.276}{.027} \\

  \midrule
  $G_{de}$=300
   & \textcolor{lightblue}{\msinline{0.529}{.131}} & \textcolor{lightblue}{\msinline{0.570}{.083}} & \textcolor{lightblue}{\msinline{0.315}{.135}}
   & \msinline{0.416}{.025} & \msinline{0.524}{.015} & \msinline{0.304}{.029} \\

  $G_{de}$=1000
   & \textcolor{midorange}{\underline{\msinline{0.517}{.154}}} & \msinline{0.578}{.085} & \msinline{0.312}{.138}
   & \textcolor{midorange}{\underline{\msinline{0.405}{.021}}} & \textcolor{midorange}{\underline{\msinline{0.511}{.012}}} & \msinline{0.304}{.027} \\

  \midrule

  Avg $\alpha_s$=0.5
   & \msinline{0.544}{.151} & \msinline{0.570}{.085} & \msinline{0.303}{.145}
   & \msinline{0.447}{.023} & \msinline{0.545}{.014} & \msinline{0.294}{.033} \\

  \underline{DUET(Ours)}
   & \textcolor{lightblue}{\msinline{0.530}{.141}} &\textcolor{midorange}{\underline{\msinline{0.567}{.081}}} & \textcolor{midorange}{\underline{\msinline{0.316}{.135}}}
   & \textcolor{lightblue}{\msinline{0.407}{.023}} & \textcolor{lightblue}{\msinline{0.520}{.013}} & \textcolor{midorange}{\underline{\msinline{0.311}{.030}}} \\

  \bottomrule
  \end{tabular}
}
\end{table*}

\noindent \textbf{Ablation Study.} We conducted a detailed ablation study with 300 HVGs to evaluate the effectiveness of each functional block, as is summarized in Table~\ref {tab:ablation_modules}. Results for DUET$_{Reg}$ and DUET$_{Ret}$ demonstrated that incorporation of cellular prior leads to clear improvement in either regression or retrieval manner. Sensitivity analysis of gating hyperparameters $\tau_c$ and $\tau_p$ further demonstrates the impact of the cellular signal. A low threshold admits many mismatched references and adds noise, whereas a high threshold discards informative signals; therefore, a moderate threshold yields the best performance. Our adaptive fusion strategy further improves performance over either single branch, reducing MSE from $0.478\pm0.031$ to $0.407\pm0.023$. The gains from each module are orthogonal, and DUET achieves optimal performance with all blocks integrated.

We further vary the size of gene panels $g_{de}$ for cell type deconvolution and observe that only a small set of prior genes is sufficient to achieve robust bio-guidance. Moreover, DUET contains 97M parameters, fewer than or similar to OmiCLIP (ViT-L/16, 304M) and UMPIRE (ViT-B/16, 87M), yet achieves superior performance, highlighting its parameter efficiency and effective architecture. 

%% file: Sec/5_conclusion.tex
\section{Conclusion}

We present a novel cellular-empowered dual-paradigm framework that moves beyond monolithic vision-to-expression modeling in ST prediction. By orchestrating parametric regression and memory-based retrieval under cellular inductive priors, DUET reconciles expressive flexibility with biological fidelity. Embedded large-scale scRNA-seq knowledge then anchors predictions to biologically grounded molecular states, promoting principled and faithful inference. An adaptive expert triage mechanism further dynamically modulates the preference of dual pathways across spatial contexts, thereby yielding a more reliable estimation. Extensive experiments on three public datasets validate our framework's superior performance against diverse prediction architectures, with consistent gains under various settings. By bridging expressive modeling with molecular reliability, DUET establishes a unified paradigm for spatial molecular modeling and paves the way toward scalable deployment in real-world omics analysis.